\documentclass[letterpaper]{article}
\usepackage{arxiv}
\usepackage{times}
\usepackage{helvet}
\usepackage{courier}
\usepackage{url}
\usepackage{graphicx}
\usepackage{xcolor}

\title{An Evaluation of Classification and Outlier Detection Algorithms\thanks{This work is supported
by InnovateUK (Grant 103682) and Digital Creativity Labs jointly funded by EPSRC/AHRC/InnovateUK grant EP/M023265/1.}}
\author{Victoria J. Hodge \and Jim Austin\\
Digital Creativity Labs, Department of Computer Science, University of York, UK\\
\{victoria.hodge, jim.austin\}@york.ac.uk
}
\begin{document}
\maketitle

\begin{abstract}
This paper evaluates algorithms for classification and outlier detection accuracies in temporal data. We focus on algorithms that train and classify rapidly and can be used for systems that need to incorporate new data regularly. Hence, we compare the accuracy of six fast algorithms using a range of well-known time-series datasets. The analyses demonstrate that the choice of algorithm is task and data specific but that we can derive heuristics for choosing. Gradient Boosting Machines are generally best for classification but there is no single winner for outlier detection though Gradient Boosting Machines (again) and Random Forest are better. Hence, we recommend running evaluations of a number of algorithms using our heuristics.

\end{abstract}
\section{Introduction}\label{sec:intro}
We analyze a range of algorithms used for both  classification and outlier detection \cite{hodge2004survey} on multi-variate time-series data \cite{keogh2003need}. Our research question is to identify an efficient method for on-line classification and outlier detection in multivariate sensor data across IoT and transport domains \cite{hodge2015wireless}. We focus on algorithms that train and classify in less than 5 minutes for all datasets here and, hence, can be used in an on-line system where new data need to be assimilated constantly. 
Our ongoing research will have labeled training data allowing supervised training. Accuracy is our key metric as we want to ensure that all anomalies are found and that classification accuracy is as high as possible. Precision is less important for our current work as false positives can be post-processed.
\\
\\
Authors have evaluated classification \cite{bagnall_2017,brown2012experimental,demvsar2006statistical} and outlier detection \cite{swersky2016evaluation} algorithms across a range of datasets. Bagnall et al. \cite{bagnall_2017} provide a comprehensive survey of time-series classification algorithms. However, none of these papers provides a general, unbiased evaluation of multi-variate time-series data for both classification and supervised outlier detection. Keogh and Kasetty \cite{keogh2003need} identified that such evaluations are vital. Additionally, as noted by Bagnall et al. \cite{bagnall_2017}, in previous classification evaluations ``several different programming languages have been used, experiments involved a
single train/test split and some used normalized data whilst others did not'' also ``the majority of the datasets were synthetic and created by the proposing authors, for the purpose of showcasing their algorithm''. 

\section{Algorithms}\label{sec:algos}
To ensure consistency of evaluation, we take as many algorithms as possible from a single framework, WEKA \cite{Weka2016}. We evaluate five popular WEKA algorithms: (Bayesian Network (\textbf{Bayes}), \textbf{C4.5}, \textbf{k-NN}, Naive Bayes (\textbf{Naive}) and Random Forest (\textbf{RF})) that are used widely. We also include Microsoft Light Gradient Boosting Machine (\textbf{GBM}) \cite{Microsoft:2016} as GBM algorithms have performed consistently well in recent Kaggle competitions for classifying and detecting anomalies in time-series data, for example \cite{taieb2014gradient}. We also evaluated WEKA logistic regression (LR)  but results were poor and are not included. 
\section{Datasets}\label{sec:data}
There are many data sets available for classification evaluations, see \cite{bagnall_2017} for a list. However, there are very few time-series datasets amenable to both outlier detection and classification evaluation. The data set should have multiple data attributes, amenable to generating time-series windows, contain at least 2000 examples, and have a clear outlier class. We scanned as many repositories as we could find and were able to extract 10 time-series datasets that met our criteria for both classification and outlier detection. The 10 datasets are: Electric Devices (\textbf{Elec}) and Ford Classification Challenge using clean and noisy data (\textbf{Ford1} and \textbf{Ford2}) respectively from the UCR data repository \cite{UCRArchive} and EEG Eye State (\textbf{EEG}); Human Activity Recognition Using Smartphones (\textbf{HAR}); five people performing different activities (\textbf{JSI}); Ozone Level Detection over 1 hour ($\mathbf{O_3^{1hr}}$) and 8 hours ($\mathbf{O_3^{8hr}}$); Occupancy Detection (\textbf{Occ}) and sensor readings of the Pioneer-1 mobile robot (\textbf{Robot}) from UCI machine learning repository \cite{Lichman:2013}. \begin{table*}
\centering
\caption{Details of the 10 datasets: training size, test size, number of attributes and outlier class. An asterisk indicates that we buffered the data into time-series length 20.}
\label{tab:1}       
\begin{tabular}{lrrrrrrrrrr}
\hline\noalign{\smallskip}
Dataset &
JSI &
EEG &
$Ford_1$ &
$Ford_2$ &
$O_3^{1hr}$ &
$O_3^{8hr}$ &
Robot &
Har &
Occ &
Elec 
  \\
\noalign{\smallskip}\hline\noalign{\smallskip}
Train &
23815 &
9874 &
3271 &
3306 &
1674 &
1672 &
4033 &
4852 &
8124 &
8926 
 \\
Test &
12268 &
5087 &
1320 &
810 &
862 &
862 &
2077 &
2500 &
9733 &
7711 
 \\
 Attr &
 $3^*$ &
$14^*$ &
500 &
500 &
72 &
72 &
$35^*$ &
561 &
$5^*$ &
96
\\
Outlier &
Walking &
One &
Plus1 &
Plus1 &
Ozone &
Ozone &
Retreat &
Laying &
1 &
6
\\
\noalign{\smallskip}\hline
\end{tabular}
\end{table*}The details of the 10 datasets are given in table \ref{tab:1}: the size of the training set, the size of the test set, the number of attributes for each datum and the outlier class. 
\\
\\
Where a dataset has a specified train/test sets we used those, otherwise we used the first 66\% of the data (sorted in chronological order) as the training set and the remainder for test. This ensures that the training data all precede the test data and we are not using future data to predict past data. The data were not cleaned nor missing values imputed as all algorithms handle missing values. Where a dataset already contains the time-series we used that; otherwise we used a time-series window of 20 points which replicates the data length in our IoT/transport problems. The vector $X_j$ represents a time-series. We slide a window representing a preset time interval over $X_j$ to subdivide it into buffers $X_j^{TS}$. Buffering always preserves the temporal ordering of the data. 
\\
$X_j^{TS}$ is \{$x_{1_{t-19}}$, $x_{1_{t-18}}$, $x_{1_{t-17}}$, ... ,$x_{y_{t-2}}$, $x_{y_{t-1}}$, $x_{y_t}$\} for 20 time slices \{$t$-19, $t$-18, ..., $t$-1, $t$\} and $y$ sensors. 
\section{Results}\label{sec:results}
The first analysis is to compare the classification accuracy of the six algorithms. We varied the parameter settings of the various algorithms and ensured that we tried an equivalent number of parameter configurations for each algorithm for equality. The accuracy is calculated across all records and classes in each dataset. \begin{table*}
\centering
\caption{Classification accuracy (\%) of the six algorithms over the 10 datasets.}
\label{tab:2}       
\begin{tabular}{lrrrrrrrrrr}
\hline\noalign{\smallskip}
Dataset &
JSI &
EEG &
$Ford_1$ &
$Ford_2$ &
$O_3^{1hr}$ &
$O_3^{8hr}$ &
Robot &
Har &
Occ &
Elec 
  \\
\noalign{\smallskip}\hline\noalign{\smallskip}
Bayes &
61.7 &
71.4 &
69.6 &
58.5 &
78.5 &
72.0 &
95.6 &
86.6 &
86.3 &
46.5
\\
C4.5 &
69.8 &
77.2 &
72.0 &
58.3 &
98.7 &
94.8 &
98.5 &
82.0 &
96.4 &
56.0
\\
kNN &
72.8 &
\textbf{79.5} &
68.7 &
60.1 &
98.6 &
95.5 &
85.5 &
91.5 &
92.9 &
59.3
\\
Naive &
54.7 &
22.8 &
53.7 &
\textbf{62.1} &
66.9 &
64.4 &
83.3 &
79.9 &
95.6 &
51.3 
\\
RF &
70.8 &
60.3 &
86.1 &
60.6 &
98.7 &
96.1 &
98.9 &
92.6 &
\textbf{98.0} &
65.5 
\\
GBM &
\textbf{75.3} &
55.8 &
\textbf{88.6} &
60.2 &
\textbf{98.8} &
\textbf{98.7} &
\textbf{99.4} &
\textbf{94.0} &
96.2 &
\textbf{67.4} 
\\
\noalign{\smallskip}\hline
\end{tabular}
\end{table*}The results are listed in table \ref{tab:2}. 
\\
\\
The second analysis compares the outlier detection accuracy of the six algorithms. Again, we varied the parameter settings of the algorithms and evaluated an equivalent number of configurations. For outlier detection, some datasets have an outlier class which we used. Otherwise, we calculated the accuracy using  the class most dissimilar to the others, for example, a walking activity when all other activities are sedentary (sitting, lying, standing, etc.); see table \ref{tab:1} for details of the data sets and which class we used as the outlier class. The accuracy results for the data sets are listed in table \ref{tab:3}.\begin{table*}
\centering
\caption{Outlier detection accuracy (\%) of the six algorithms over the 10 datasets.}
\label{tab:3}       
\begin{tabular}{lrrrrrrrrrr}
\hline\noalign{\smallskip}
Dataset &
JSI &
EEG &
$Ford_1$ &
$Ford_2$ &
$O_3^{1hr}$ &
$O_3^{8hr}$ &
Robot &
Har &
Occ &
Elec 
  \\
\noalign{\smallskip}\hline\noalign{\smallskip}
Bayes &
78.3 &
69.4 &
80.9 &
95.6 &
81.8 &
82.4 &
92.3 &
98.8 &
90.8 &
18.3
\\
C4.5 &
85.1 &
0.0 &
70.3 &
76.5 &
18.2 &
26.5 &
95.4 &
\textbf{100.0} &
84.1 &
\textbf{39.0}
\\
kNN &
89.7 &
19.8 &
43.0 &
48.4 &
27.3 &
35.3 &
70.3 &
97.3 &
80.9 &
28.3
\\
Naive &
65.5 &
58.2 &
24.9 &
90.7 &
\textbf{90.9} &
\textbf{91.2} &
84.1 &
94.4 &
88.0 &
37.8
\\
RF &
89.9 &
\textbf{75.8} &
\textbf{94.5} &
\textbf{98.5} &
0.0 &
0.0 &
97.1 &
97.5 &
\textbf{92.2} &
32.2 
\\
GBM &
\textbf{93.3} &
28.5 &
92.6 &
81.8 &
9.1 &
0.0 &
\textbf{97.3} &
\textbf{100.0} &
91.5 &
21.4 
\\
\noalign{\smallskip}\hline
\end{tabular}
\end{table*}\section{Analysis}
For classification (see table \ref{tab:2}), GBM outperformed all other algorithms across all datasets. It achieves the highest accuracy on 7 of the 10 datasets. It only underperformed on one dataset: EEG Eye state where GBM achieves 55.8\% accuracy compared to 79.5\% accuracy for k-NN. The fact that an instance-based learner outperformed compared to the model-based approaches indicates that this dataset is difficult to model with a complex cluster structure. This is supported by checking with linear regression which only achieved 28.5\% accuracy so the problem is highly non-linear. Additionally, C4.5 (one decision tree) significantly outperformed RF and GBM (ensembles of trees) on the EEG Eye state data set indicating that an exact fit to the data is required and an ensemble approach will not be specific enough to be accurate. 
\\
\\For classification on these time-series data, GBM is the best method overall from a general perspective but cannot be guaranteed to be the best so other algorithms need to be considered.
\\
\\
For outlier detection (see table \ref{tab:3}), the picture is much less clear. Bayes had the highest average accuracy across all datasets but is not best on any dataset. RF is best on 4 datasets but does not find any outliers in the two Ozone datasets. GBM is best on 3 datasets but also struggled with the two Ozone datasets. These two datasets are very unbalanced with only 7.5\% outliers and 92.5\% normal. In contrast, both RF and GBM performed well for classification on these data sets. They are classifying normal data well but misclassifying outliers suggesting over-fitting. C4.5 and Naive were best on 2 datasets. Similarly, C4.5 under-performs for outlier detection on data sets it classified well indicating over-fitting. Naive seems less prone to over-fitting and tends to perform poorly for outlier detection on the same data sets it performed poorly on for classification.
\\
\\For outlier detection, we recommend considering a number of algorithms. We recommend GBM and RF for high accuracy on some data sets but poor on unbalanced data. The data set can be pre-processed to balance it if sufficient data are available or a cost matrix used with the algorithm to penalize misclassification of outliers or the tree size limited to prevent over-fitting. Similarly C4.5 can perform well but over-fits some data sets. Naive can perform well but struggle with some data sets. Bayes is good on most data sets but never the best. The only algorithms that can be discounted for outlier detection on these data are LR and k-NN which had low overall accuracy and do not perform best on any datasets.  
\section{Conclusion}
In this paper, we evaluated algorithms for both classification and outlier detection for an on-line system that assimilates new data regularly. We aimed to derive heuristics for the best algorithms. For classification, GBM was best. However, it did not excel on every dataset. For a dataset with complex structure (non-linearity) where an ensemble method will fail and a very specific approach is needed, then k-NN or C4.5 are better. It is not possible to derive a simple heuristic for outlier detection indicating that a number of algorithms need to be evaluated. We ruled out LR and k-NN as they underperformed across all datasets. Bayes is a good all round algorithm but does not excel. Hence, RF and GBM are generally better then C4.5 or Naive so these are the likely algorithms to try. C4.5, GBM and RF struggled on datasets with very few outlier examples in the training data probably due to over-fitting. It may be possible to overcome over-fitting by introducing a cost matrix or limiting tree size. Alternatively, if sufficient data are available then the data set may be balanced.
\\
\\
As with \cite{bagnall_2017}, we have not exhaustively optimized the algorithms under investigation. For example, we fixed the maximum number of parameter combinations tried for each algorithm rather than fixing the maximum processing time for each algorithm. This could be considered unfair to faster algorithms which could evaluate more parameter combinations in a fixed time but we felt provides more fairness for an accuracy evaluation.  

\bibliography{refs}
\bibliographystyle{arxiv}

\end{document}